\documentclass[runningheads]{llncs}

\usepackage[T1]{fontenc}
\usepackage{graphicx}
\usepackage{verbatim}
\usepackage{subcaption}
\usepackage{placeins}
\usepackage[hidelinks,breaklinks=true]{hyperref}

\begin{document}

\title{3D Digital Twin Visualization of Multiclass GRF-Based Gait Disorder Classification}
\titlerunning{3D Digital Twin Visualization of GRF-Based Gait Disorder Classification}

\author{Nayoung Son\inst{1}
\and
Minwoo Shin\inst{1}\thanks{Corresponding author.}
}


\authorrunning{N. Son and M. Shin}

\institute{
Department of Software, Yonsei University, Wonju, 26493, Republic of Korea\\
\email{mshin@yonsei.ac.kr}
}

\maketitle

\begin{abstract}
Automated gait analysis requires accurate classification and interpretable
outputs. We propose an integrated framework for classifying healthy gait
and multiple musculoskeletal impairment groups using bilateral ground
reaction force (GRF) and center-of-pressure (COP) signals. The signals
were normalized over the stance phase and standardized using training-set
statistics. The model achieved a validation accuracy of 99.00\% and a test
accuracy of 90.07\% under a session-level split. Class-specific
$\epsilon$-LRP identified positive and negative contributions across both
sides, multiple signal components, and different stance phases. Separately,
the processed GRF signals and model predictions were synchronized within
a Blender-based 3D visualization, enabling sample-level inspection of gait
trials and classification results. The proposed framework integrates
classification, explainability, and 3D visualization to improve model
transparency. The source code is available in the following repository: \url{https://github.com/nyoico/grf-gait-3d-visualization.git}

\keywords{Gait Disorder \and Ground Reaction Force \and Multiclass Classification \and Transformer \and Explainable AI \and 3D Visualization}
\end{abstract}

\section{Introduction}
Instrumented gait analysis provides objective measurements for assessing human locomotion and identifying pathological gait patterns~\cite{baker2006gait,whittle1996clinical,muro2014gait}. Among biomechanical signals, ground reaction force (GRF) is widely used because it reflects weight acceptance, propulsion, balance, and limb loading during walking~\cite{horsak2020gaitrec,park2022grf}. However, clinical GRF recordings are high-dimensional, temporally dependent, and variable across subjects and impairment types, making manual interpretation difficult and motivating automated gait classification systems~\cite{chau2001review1,chau2001review2,figueiredo2018automatic,khera2020role,seo2026acoustic,cho2026dt}.

Machine learning has been widely explored for gait disorder classification using GRF and related biomechanical measurements~\cite{figueiredo2018automatic,prakash2018recent,khera2020role,harris2022survey}. Conventional approaches, including support vector machines, nearest-neighbor classifiers, self-organizing neural networks, and other analytical models, have been used to categorize gait patterns into clinically meaningful groups~\cite{begg2005machine,mezghani2008automatic,barton2007gait}. Large-scale annotated datasets such as GaitRec have further enabled data-driven modeling of healthy and impaired gait, including musculoskeletal impairments involving the hip, knee, ankle, and calcaneus~\cite{horsak2020gaitrec}. Related public GRF resources, such as the Gutenberg Gait Database, also support reproducible gait modeling and benchmarking~\cite{horst2021gutenberg}. Nevertheless, two limitations remain important. First, many classification models provide limited interpretability, making it difficult to understand which gait phases or force components drive the decision~\cite{dindorf2020interpretability}. Second, model outputs are often presented only as numerical metrics or plots, limiting intuitive inspection of individual gait trials; visual analytics and digital twin approaches provide a complementary route for sample-level interpretation~\cite{rind2022trustworthy,katsoulakis2024digital,uhlenberg2023cosimulation}.

To address these limitations, we propose an integrated framework for interpretable bilateral gait classification and digital twin-based visualization. The proposed model uses an encoder--decoder Transformer architecture to classify musculoskeletal gait patterns from left and right GRF/COP signals, motivated by the success of attention-based models in sequence and multivariate time-series representation learning~\cite{fawaz2019deep}. To improve transparency, we apply class-specific $\epsilon$-LRP to identify temporally localized and channel-wise relevance patterns across the stance phase~\cite{bach2015pixel,samek2017evaluating}. In addition, we develop a Blender-based gait digital twin simulation that synchronizes GRF values, ground-truth labels, and predicted labels within a 3D virtual gait environment~\cite{wickes2009blender,fuller2020digital,jones2020characterising}. This enables both quantitative evaluation and sample-level qualitative inspection of model predictions.

The main contributions of this study are as follows. First, we develop a bilateral Transformer-based classifier for multiclass musculoskeletal gait classification using time-normalized GRF/COP signals. Second, we introduce class-specific $\epsilon$-LRP analysis to visualize positive and negative relevance across stance phases and force components. Third, we integrate the model outputs into a Blender-based data-driven gait digital twin, allowing individual test samples to be examined together with synchronized biomechanical signals and classification results.

\section{Method}
\subsection{Data and Input Definition}
Raw GRF and COP data were preprocessed at the session and trial level. Each gait trial was represented as 101 time-normalized stance-phase points. For each side, five channels---COP\_AP, COP\_ML, F\_AP, F\_ML, and F\_V---were aligned using SESSION\_ID and TRIAL\_ID, and only trials common to all channels were retained. Here, COP denotes the center of pressure, while AP, ML, and V indicate the anterior--posterior, mediolateral, and vertical directions, respectively. Therefore, COP\_AP and COP\_ML represent the center-of-pressure trajectories in the anterior--posterior and mediolateral directions, while F\_AP, F\_ML, and F\_V represent the ground reaction force components along the anterior--posterior, mediolateral, and vertical axes.

SESSION\_ID denotes the unique identifier of each recording session, and TRIAL\_ID indicates the repeated trial index within a session. Diagnostic labels were assigned from CLASS\_LABEL\_DETAILED metadata using SESSION\_ID, where CLASS\_LABEL\_DETAILED represents the detailed diagnostic class of each session. The classifier used 12 output labels, including healthy controls and
musculoskeletal impairment groups. Table~\ref{tab1} summarizes
the class definitions and their use across the data splits. Because the
C\_A class contained only one sample in the entire dataset, it was
retained only in the training set and excluded from validation and test
evaluation. For the remaining 11 classes, a stratified split was
performed at the session level to prevent data leakage among the
training, validation, and test sets. Left and right side data were saved separately as arrays with shape $N \times 101 \times 5$, where $N$ denotes the number of gait trials.

\begin{table}[t]
\centering
\caption{Classification labels and evaluation usage.}\label{tab1}
{\fontsize{8pt}{9.5pt}\selectfont
\begin{tabular}{|l|l|l|l|}
\hline
Category & Class Name & Class Definition & Use\\
\hline
Healthy & HC & Healthy control & Train/Test\\
Hip & H\_P & Pelvis & Train/Test\\
Hip & H\_C & Coxa & Train/Test\\
Hip & H\_F & Femur & Train/Test\\
Knee & K\_P & Patella & Train/Test\\
Knee & K\_F & Knee-region fracture & Train/Test\\
Knee & K\_R & Rupture of ligaments or menisci & Train/Test\\
Ankle & A\_F & Ankle-region fracture & Train/Test\\
Ankle & A\_R & Rupture of ligaments or Achilles tendon & Train/Test\\
Ankle & A\_L & Lower-leg shaft fracture & Train/Test\\
Calcaneus & C\_F & Fracture & Train/Test\\
Calcaneus & C\_A & Arthrodesis & Train only\\
\hline
\end{tabular}
}
\end{table}

\subsection{Network}
\begin{figure}[t!]
\centering
\includegraphics[width=0.6\textwidth]{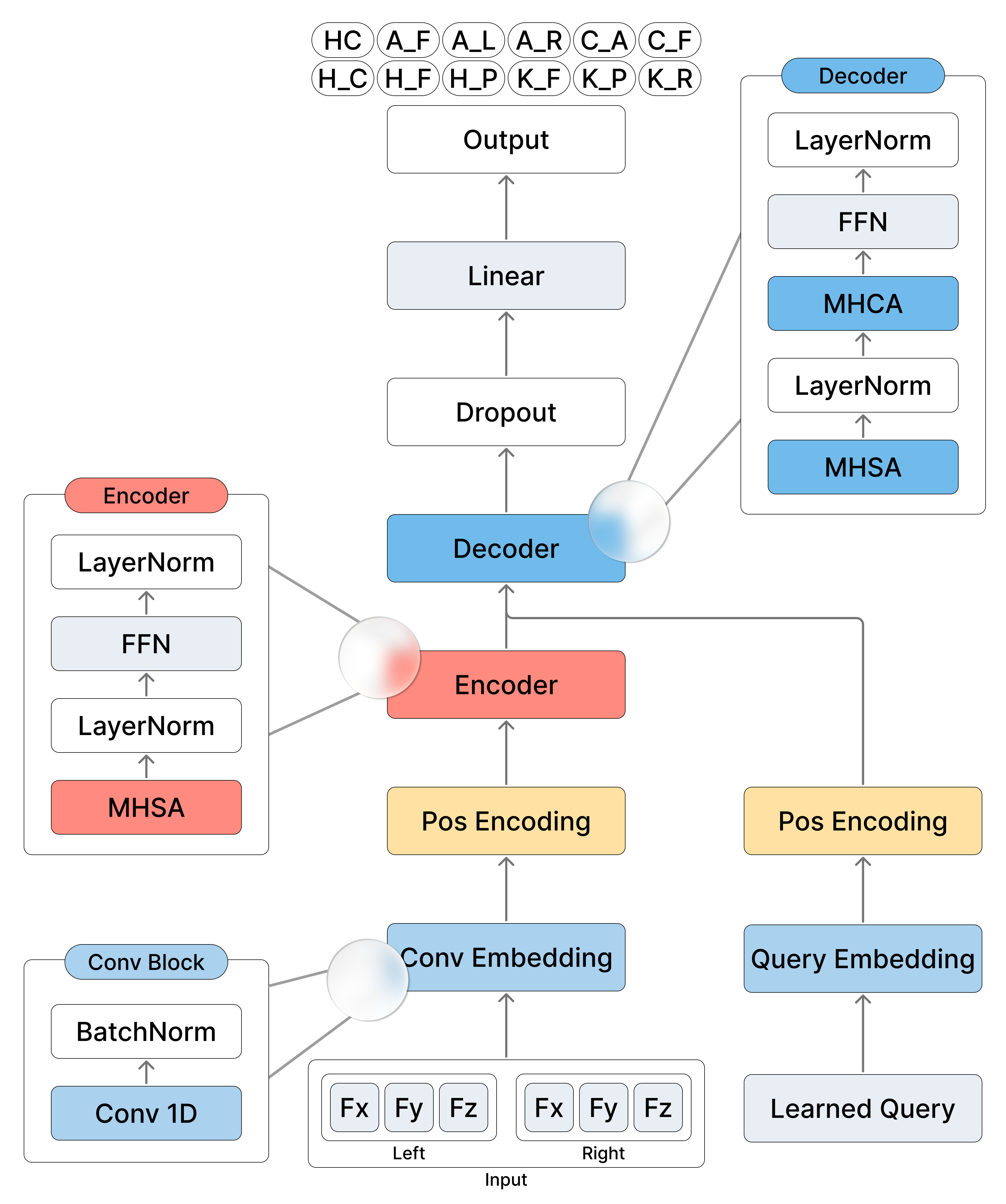}
\caption{Overall architecture of the proposed Transformer encoder--decoder classifier. Bilateral COP and GRF sequences are processed through side-specific convolutional embeddings and positional encoding. The encoder models temporal gait representations, and a learned decoder query attends to the encoder memory to produce 12 class logits.}\label{fig1}
\end{figure}

The proposed model uses an encoder--decoder Transformer for bilateral gait classification, as illustrated in Fig.~\ref{fig1}. Left and right GRF/COP sequences are first projected using side-specific 1D convolutional embeddings and combined with sinusoidal positional encoding, thereby preserving bilateral input characteristics before temporal modeling. The Transformer encoder models temporal dependencies across the 101-point stance phase using multi-head self-attention, while a learned decoder query attends to the encoded gait representation through cross-attention. As shown in Fig.~\ref{fig1}, the resulting decoder feature is passed to dropout and a linear classifier to produce 12 class logits. The main architectural settings are summarized in Table~\ref{tab2}.

\subsection{Architecture-Aware $\epsilon$-LRP for Gait Model Explanation}
Class-specific relevance propagation was performed using each output logit as the initial relevance score. Relevance was propagated backward through the classifier, decoder, encoder, and convolutional embedding layers to obtain time- and channel-wise relevance scores for signals from both sides. For class-level analysis, relevance scores from all test samples within the same class were averaged to reduce trial-level variability.

The implemented method uses customized $\epsilon$-LRP rules adapted to the proposed architecture. Epsilon-stabilized rules were applied to the linear classifier and convolutional layers, feed-forward networks were handled through their Conv1D--ReLU--Conv1D structure, and attention modules were treated with a pass-through strategy. Because the left and right embeddings are combined by summation, relevance was equally distributed to both branches and propagated back to the input channels. Signed relevance was visualized over ML, AP, and vertical force components: positive relevance indicates evidence supporting the target class logit, whereas negative relevance indicates evidence suppressing it or separating it from competing classes. These relevance maps were interpreted as model-derived evidence rather than direct clinical causality.

\subsection{3D Gait Visualization in Blender}
A Blender-based visualization module was developed to inspect individual test trials in a virtual gait analysis scene. The module reads the inference CSV containing true labels, predicted labels, and bilateral GRF time series, then updates the displayed GRF values according to the current stance-phase frame. The scene includes a humanoid model, bilateral force plates, and a text panel showing the sample index, time index, GRF components, true class, and predicted class. This provides a sample-level interface for jointly reviewing gait signals and model predictions.

\begin{figure}[t]
\centering
\includegraphics[width=0.7\textwidth]{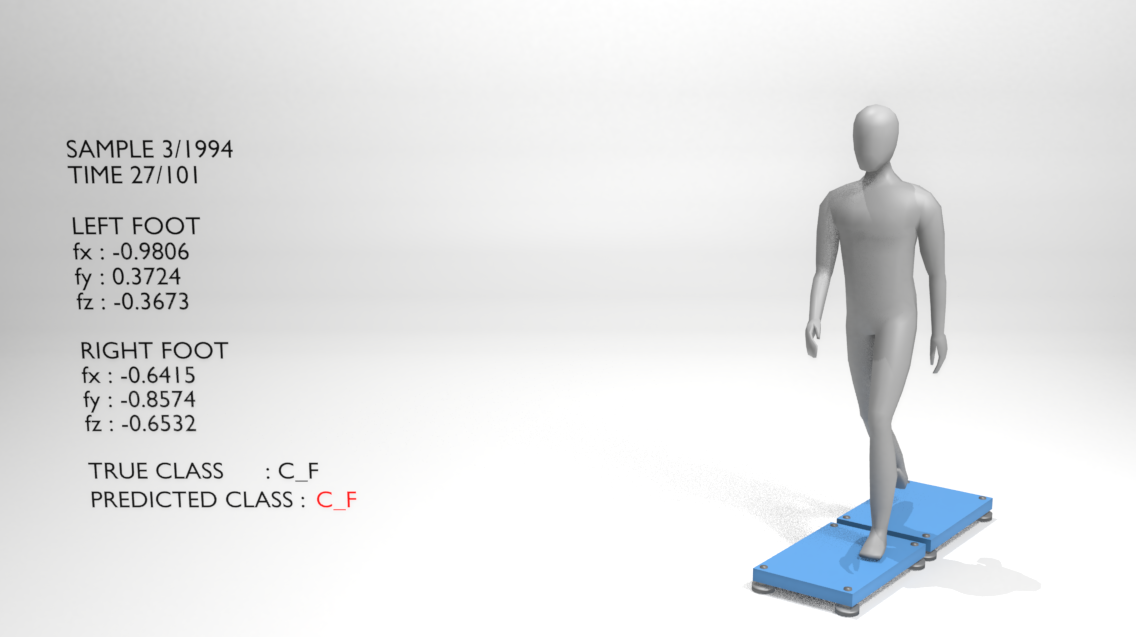}
\caption{Blender-based 3D visualization of the gait classification result. A humanoid model is placed on bilateral force plates, while the text panel displays the current sample index, stance-phase time index, bilateral GRF values, true class, and predicted class.}\label{fig2}
\end{figure}

\section{Experiments}
\subsection{Dataset and Evaluation Protocol}
Experiments were conducted on the post-processed GaitRec dataset using the 12 target classes in Table~\ref{tab1}. The session-level split produced 61,266 training, 1,894 validation, and 1,994 independent test samples. Table~\ref{tab3} summarizes the training hyperparameters and inference configuration used in the experiments, including the data split setting, optimization parameters, learning-rate scheduler, early stopping criterion, and latency measurement protocol. The model was trained using cross-entropy loss with label smoothing, AdamW optimization, ReduceLROnPlateau scheduling, and early stopping based on validation performance. The best validation checkpoint was used for final testing. Performance was evaluated using accuracy, confusion matrix, class-wise precision/recall/F1-score, and single-sample inference latency after warm-up iterations.

\begin{table}[t]
\centering
\caption{Model configuration of the proposed Transformer-based gait classification network.}\label{tab2}
{\fontsize{8pt}{9.5pt}\selectfont
\begin{tabular}{|l|l|l|}
\hline
Components & Setting & Value\\
\hline
Input sequence length & Stance-normalized time points & 101\\
Output classes & Gait condition class & 12\\
Embedding & Embedding dimension & 256\\
Transformer Encoder--Decoder & Heads / layers & 8 / 4\\
Feed-forward network & Hidden dimension & 1024\\
Regularization & Dropout rate & 0.1\\
Model size & Total parameters & 7.38M\\
\hline
\end{tabular}
}
\end{table}

\begin{table}[t]
\centering
\caption{Training hyperparameters and inference configuration. Latency was measured on CUDA with batch size 1 after 20 warm-up iterations and averaged over 100 repeats.}\label{tab3}
{\fontsize{8pt}{9.5pt}\selectfont
\begin{tabular}{|l|l|l|}
\hline
Category & Setting & Value\\
\hline
Data split & Random seed / Validation ratio & 42 / 0.03\\
Training & Batch size / Max epochs & 32 / 300\\
 & Early stopping patience & 20\\
Optimization & Optimizer & AdamW\\
 & Learning rate / Weight decay & $3 \times 10^{-4}$ / $1 \times 10^{-4}$\\
 & Loss / Label smoothing & Cross-entropy / 0.05\\
Scheduler & Type / Mode & ReduceLROnPlateau / max\\
 & Factor / Patience / MinLR & 0.5 / 5 / $1 \times 10^{-6}$\\
Inference & Batch size / Device & 1 / CUDA\\
 & Warm-up / Repeats & 20 / 100\\
 & Mean latency & 4.89 ms\\
Inference Hardware & GPU model & NVIDIA GeForce RTX 5080 \\
\hline
\end{tabular}
}
\end{table}

\subsection{Quantitative Classification Results}
As summarized in Table~\ref{tab4}, the model achieved 98.997\% validation accuracy at epoch 97 and 90.07\% test accuracy on 1,994 independent test samples.

Table~\ref{tab5} reports the class-wise recall on the independent test set. High recall was observed for several classes, including C\_F, H\_F, K\_F, and HC, whereas relatively lower recall was observed for K\_R and A\_R. These results suggest that the proposed model performed well for most gait classes, while some clinically related impairment groups remained more difficult to distinguish due to similar biomechanical characteristics. C\_A was not evaluated because it had no test samples. The final model contained 7.38M trainable parameters and achieved 4.89 $\pm$ 0.36 ms single-sample inference latency on an RTX 5080.

\begin{table}[t!]
\centering
\caption{Quantitative summary of the proposed model.}\label{tab4}
\begin{tabular}{|l|l|}
\hline
Metric & Value\\
\hline
Best epoch & 97\\
Best validation accuracy & 99.0\%\\
Test accuracy & 90.1\%\\
Total parameters & 7.38M\\
Trainable parameters & 7.38M\\
Inference device & CUDA\\
Inference latency & $4.89 \pm 0.36$ ms \\
\hline
\end{tabular}
\end{table}

\begin{table}[t!]
\centering
\caption{Class-wise recall (\%) on the independent test set. Support and correct denote total and correctly classified samples. C\_A was excluded because it was absent from the test split.}\label{tab5}
\begin{tabular}{|l|*{12}{c|}}
\hline
Class & HC & H\_P & H\_C & H\_F & K\_P & K\_F & K\_R & A\_F & A\_R & A\_L & C\_F & C\_A\\
\hline
Support & 240 & 40 & 134 & 163 & 66 & 257 & 137 & 347 & 28 & 138 & 444 & 0\\
Correct & 220 & 34 & 117 & 162 & 43 & 239 & 110 & 302 & 22 & 120 & 427 & 0\\

Recall & 91.7 & 85.0 & 87.3 & 99.4 & 65.2 & 93.0 & 80.3 & 87.0 & 78.6 & 87.0 & 96.2 & N/A\\
\hline
\end{tabular}
\end{table}

\subsection{Class-Specific $\epsilon$-LRP Analysis}
\begin{figure}[t!]
\centering
\includegraphics[width=\textwidth]{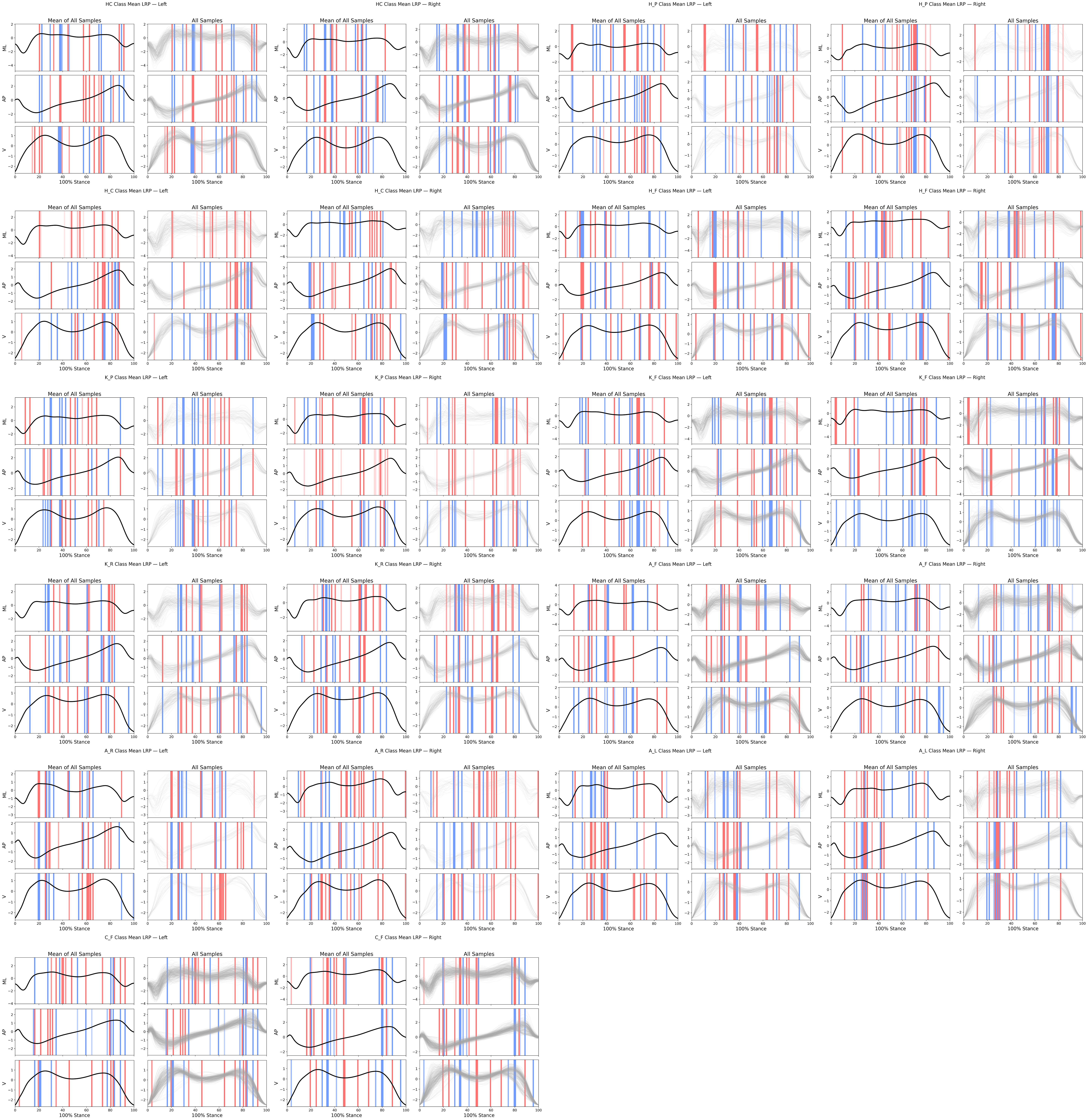}
\caption{Class-specific mean signed $\epsilon$-LRP maps for the 11 gait
classes represented in the test set. Red and blue bands indicate positive
and negative relevance, respectively, across the stance phase for the
left- and right-side ML, AP, and vertical force components.}
\label{fig5}
\end{figure}

Class-specific $\epsilon$-LRP was applied to the 11 classes present in the test split. Fig.~\ref{fig5} shows the class-specific mean signed $\epsilon$-LRP maps for the left and right sides. Each subplot visualizes the averaged GRF waveform together with relevance bands across the stance phase, where red bands indicate positive relevance supporting the corresponding class prediction and blue bands indicate negative relevance opposing the prediction. The relevance maps showed class- and side-dependent evidence across stance phases and across ML, AP, and vertical force components. In particular, the model did not rely uniformly on the entire stance phase; instead, it emphasized localized temporal regions that differed by class and side. This suggests that the proposed Transformer captured discriminative gait features related to side-specific loading, braking/propulsion, and vertical force patterns. Signed relevance indicated both supporting and opposing evidence, showing that the model used localized biomechanical patterns to separate impairment classes.

\subsection{Blender-Based Gait Digital Twin Simulation}
The Blender-based digital twin was used for qualitative inspection of individual test samples. As shown in Fig.~\ref{fig2}, the scene displays a humanoid model on bilateral force plates and an information panel with synchronized GRF values, true labels, and predicted labels. This interface complements quantitative metrics and $\epsilon$-LRP by enabling sample-specific review of gait signals and model predictions.

\FloatBarrier
\section{Conclusion}
This study proposed an integrated framework for gait classification, model interpretation, and 3D digital twin visualization using bilateral GRF/COP signals. The encoder--decoder Transformer achieved 90.07\% test accuracy with fast single-sample inference, while class-specific $\epsilon$-LRP revealed temporally localized relevance patterns across force components and sides. The Blender-based digital twin enabled sample-level inspection with synchronized GRF values, true labels, and predicted labels. Future work will incorporate richer motion dynamics, external validation, and broader clinical evaluation.

\begin{credits}
\subsubsection{\ackname}

This work was supported by the Korea Medical Device Development Foundation grant funded by the Korean government (the Ministry of Science and ICT, the Ministry of Trade, Industry and Energy, the Ministry of Health and Welfare, and the Ministry of Food and Drug Safety) (Grant No. RS-2026-25543484).

This research was also supported by the ANCHOR Program through the Gangwon ANCHOR Center, funded by the Ministry of Education (MOE) and Gangwon State (G.S.), Republic of Korea (Grant No. 2026-ANCHOR-10-006).

This research was further supported by the Ministry of Science and ICT (MSIT), Korea, under the National Program in Medical AI Semiconductor (Grant No. 2024-0-00096), supervised by the Institute of Information \& Communications Technology Planning \& Evaluation (IITP) in 2026.

\subsubsection{\discintname}
The authors have no competing interests to declare that are relevant to the content of this article.

\end{credits}

\bibliographystyle{splncs04}
\bibliography{Paper-0011}

@article{baker2006gait,
  author  = {Baker, Richard},
  title   = {Gait analysis methods in rehabilitation},
  journal = {Journal of NeuroEngineering and Rehabilitation},
  volume  = {3},
  pages   = {4},
  year    = {2006},
  doi     = {10.1186/1743-0003-3-4}
}

@article{whittle1996clinical,
  author  = {Whittle, Michael W.},
  title   = {Clinical gait analysis: A review},
  journal = {Human Movement Science},
  volume  = {15},
  number  = {3},
  pages   = {369--387},
  year    = {1996},
  doi     = {10.1016/0167-9457(96)00006-1}
}

@article{muro2014gait,
  author  = {Muro-de-la-Herran, Alvaro and Garcia-Zapirain, Begonya and Mendez-Zorrilla, Amaia},
  title   = {Gait Analysis Methods: An Overview of Wearable and Non-Wearable Systems, Highlighting Clinical Applications},
  journal = {Sensors},
  volume  = {14},
  number  = {2},
  pages   = {3362--3394},
  year    = {2014},
  doi     = {10.3390/s140203362}
}

@article{prakash2018recent,
  author  = {Prakash, Chandra and Kumar, Rajesh and Mittal, Namita},
  title   = {Recent developments in human gait research: parameters, approaches, applications, machine learning techniques, datasets and challenges},
  journal = {Artificial Intelligence Review},
  volume  = {49},
  number  = {1},
  pages   = {1--40},
  year    = {2018},
  doi     = {10.1007/s10462-016-9514-6}
}

@article{khera2020role,
  author  = {Khera, Preeti and Kumar, Neelesh},
  title   = {Role of machine learning in gait analysis: a review},
  journal = {Journal of Medical Engineering \& Technology},
  volume  = {44},
  number  = {8},
  pages   = {441--467},
  year    = {2020},
  doi     = {10.1080/03091902.2020.1822940}
}

@article{harris2022survey,
  author  = {Harris, Elsa J. and Khoo, I-Hung and Demircan, Emel},
  title   = {A Survey of Human Gait-Based Artificial Intelligence Applications},
  journal = {Frontiers in Robotics and AI},
  volume  = {8},
  pages   = {749274},
  year    = {2022},
  doi     = {10.3389/frobt.2021.749274}
}

@article{horsak2020gaitrec,
  author  = {Horsak, Brian and Slijepcevic, Djordje and Raberger, Anna-Maria and Schwab, Caterine and Worisch, Michael and Zeppelzauer, Matthias},
  title   = {{GaitRec}, a large-scale ground reaction force dataset of healthy and impaired gait},
  journal = {Scientific Data},
  volume  = {7},
  pages   = {143},
  year    = {2020},
  doi     = {10.1038/s41597-020-0481-z}
}

@article{horst2021gutenberg,
  author  = {Horst, Fabian and Slijepcevic, Djordje and Simak, Marvin and Schollhorn, Wolfgang I.},
  title   = {Gutenberg Gait Database, a ground reaction force database of level overground walking in healthy individuals},
  journal = {Scientific Data},
  volume  = {8},
  pages   = {232},
  year    = {2021},
  doi     = {10.1038/s41597-021-01014-6}
}

@article{park2022grf,
  author  = {Park, Ji Su and Kim, Choong Hyun},
  title   = {Ground-Reaction-Force-Based Gait Analysis and Its Application to Gait Disorder Assessment: New Indices for Quantifying Walking Behavior},
  journal = {Sensors},
  volume  = {22},
  number  = {19},
  pages   = {7558},
  year    = {2022},
  doi     = {10.3390/s22197558}
}

@article{chau2001review1,
  author  = {Chau, Tom},
  title   = {A review of analytical techniques for gait data. Part 1: Fuzzy, statistical and fractal methods},
  journal = {Gait \& Posture},
  volume  = {13},
  number  = {1},
  pages   = {49--66},
  year    = {2001},
  doi     = {10.1016/S0966-6362(00)00094-1}
}

@article{chau2001review2,
  author  = {Chau, Tom},
  title   = {A review of analytical techniques for gait data. Part 2: Neural network and wavelet methods},
  journal = {Gait \& Posture},
  volume  = {13},
  number  = {2},
  pages   = {102--120},
  year    = {2001},
  doi     = {10.1016/S0966-6362(00)00095-3}
}

@article{figueiredo2018automatic,
  author  = {Figueiredo, Joana and Santos, Cristina P. and Moreno, Juan C.},
  title   = {Automatic recognition of gait patterns in human motor disorders using machine learning: A review},
  journal = {Medical Engineering \& Physics},
  volume  = {53},
  pages   = {1--12},
  year    = {2018},
  doi     = {10.1016/j.medengphy.2017.12.006}
}

@article{begg2005machine,
  author  = {Begg, Rezaul and Kamruzzaman, Joarder},
  title   = {A machine learning approach for automated recognition of movement patterns using basic, kinetic and kinematic gait data},
  journal = {Journal of Biomechanics},
  volume  = {38},
  number  = {3},
  pages   = {401--408},
  year    = {2005},
  doi     = {10.1016/j.jbiomech.2004.05.002}
}

@article{mezghani2008automatic,
  author  = {Mezghani, Neila and Husse, Sabine and Boivin, Karine and Turcot, Katia and Aissaoui, Rachid and Hagemeister, Nicola and de Guise, Jacques A.},
  title   = {Automatic Classification of Asymptomatic and Osteoarthritis Knee Gait Patterns Using Kinematic Data Features and the Nearest Neighbor Classifier},
  journal = {IEEE Transactions on Biomedical Engineering},
  volume  = {55},
  number  = {3},
  pages   = {1230--1232},
  year    = {2008},
  doi     = {10.1109/TBME.2007.905388}
}

@article{barton2007gait,
  author  = {Barton, Gabor J. and Lisboa, Paulo J. G. and Lees, Adrian and Attfield, Steve},
  title   = {Gait quality assessment using self-organising artificial neural networks},
  journal = {Gait \& Posture},
  volume  = {25},
  number  = {3},
  pages   = {374--379},
  year    = {2007},
  doi     = {10.1016/j.gaitpost.2006.05.003}
}

@article{dindorf2020interpretability,
  author  = {Dindorf, Carlo and Teufl, Wolfgang and Taetz, Bertram and Bleser, Gabriele and Frohlich, Michael},
  title   = {Interpretability of Input Representations for Gait Classification in Patients after Total Hip Arthroplasty},
  journal = {Sensors},
  volume  = {20},
  number  = {16},
  pages   = {4385},
  year    = {2020},
  doi     = {10.3390/s20164385}
}

@inproceedings{rind2022trustworthy,
  author    = {Rind, Alexander and Slijepcevic, Djordje and Zeppelzauer, Matthias and Unglaube, Fabian and Kranzl, Andreas and Horsak, Brian},
  title     = {Trustworthy Visual Analytics in Clinical Gait Analysis: A Case Study for Patients with Cerebral Palsy},
  booktitle = {2022 IEEE Workshop on TRust and EXpertise in Visual Analytics (TREX)},
  pages     = {8--15},
  year      = {2022},
  doi       = {10.1109/TREX57753.2022.00006}
}

@article{fawaz2019deep,
  author  = {Ismail Fawaz, Hassan and Forestier, Germain and Weber, Jonathan and Idoumghar, Lhassane and Muller, Pierre-Alain},
  title   = {Deep learning for time series classification: a review},
  journal = {Data Mining and Knowledge Discovery},
  volume  = {33},
  number  = {4},
  pages   = {917--963},
  year    = {2019},
  doi     = {10.1007/s10618-019-00619-1}
}

@article{bach2015pixel,
  author  = {Bach, Sebastian and Binder, Alexander and Montavon, Gregoire and Klauschen, Frederick and Muller, Klaus-Robert and Samek, Wojciech},
  title   = {On Pixel-Wise Explanations for Non-Linear Classifier Decisions by Layer-Wise Relevance Propagation},
  journal = {PLOS ONE},
  volume  = {10},
  number  = {7},
  pages   = {e0130140},
  year    = {2015},
  doi     = {10.1371/journal.pone.0130140}
}

@article{samek2017evaluating,
  author  = {Samek, Wojciech and Binder, Alexander and Montavon, Gregoire and Lapuschkin, Sebastian and Muller, Klaus-Robert},
  title   = {Evaluating the Visualization of What a Deep Neural Network Has Learned},
  journal = {IEEE Transactions on Neural Networks and Learning Systems},
  volume  = {28},
  number  = {11},
  pages   = {2660--2673},
  year    = {2017},
  doi     = {10.1109/TNNLS.2016.2599820}
}

@article{katsoulakis2024digital,
  author  = {Katsoulakis, Evangelia and Wang, Qi and Wu, Huanmei and Shahriyari, Leili and Fletcher, Richard and Liu, Jing and Achenie, Luke and Liu, Huan and Jackson, Paul and Xiao, Yu and Syeda-Mahmood, Tanveer and Tuli, Rajeev and Deng, Jessica},
  title   = {Digital twins for health: a scoping review},
  journal = {npj Digital Medicine},
  volume  = {7},
  pages   = {77},
  year    = {2024},
  doi     = {10.1038/s41746-024-01073-0}
}

@article{uhlenberg2023cosimulation,
  author  = {Uhlenberg, Lena and Derungs, Adrian and Amft, Oliver},
  title   = {Co-simulation of human digital twins and wearable inertial sensors to analyse gait event estimation},
  journal = {Frontiers in Bioengineering and Biotechnology},
  volume  = {11},
  pages   = {1104000},
  year    = {2023},
  doi     = {10.3389/fbioe.2023.1104000}
}

@article{fuller2020digital,
  author  = {Fuller, Aidan and Fan, Zhong and Day, Charles and Barlow, Chris},
  title   = {Digital Twin: Enabling Technologies, Challenges and Open Research},
  journal = {IEEE Access},
  volume  = {8},
  pages   = {108952--108971},
  year    = {2020},
  doi     = {10.1109/ACCESS.2020.2998358}
}

@article{jones2020characterising,
  author  = {Jones, David and Snider, Chris and Nassehi, Aydin and Yon, Jason and Hicks, Ben},
  title   = {Characterising the Digital Twin: A systematic literature review},
  journal = {CIRP Journal of Manufacturing Science and Technology},
  volume  = {29},
  pages   = {36--52},
  year    = {2020},
  doi     = {10.1016/j.cirpj.2020.02.002}
}

@incollection{wickes2009blender,
  author    = {Wickes, Roger D.},
  title     = {Blender Overview},
  booktitle = {Foundation Blender Compositing},
  publisher = {Apress},
  address   = {Berkeley, CA},
  pages     = {1--23},
  year      = {2009},
  doi       = {10.1007/978-1-4302-1977-4_1}
}

@InProceedings{seo2026acoustic,
author="Seo, Minjee
and Shin, Minwoo
and Noh, Gunwoo
and Yoo, Seung-Schik
and Yoon, Kyungho",
editor="Li, Lei
and Jirsa, Viktor
and Feng, Jianfeng
and Deng, Jun
and Dede', Luca
and An, Sora
and Lyu, Yilin
and Liu, Xiaoyue",
title="Acoustic Simulation with Deep Learning for Low-Intensity Transcranial Focused Ultrasound Digital Twins",
booktitle="Digital Twin for Healthcare",
year="2026",
publisher="Springer Nature Switzerland",
address="Cham",
pages="58--68",
isbn="978-3-032-07694-6"
}

@InProceedings{cho2026dt,
author="Cho, Seonaeng
and Seo, Minjee
and Shin, Minwoo
and Yoon, Kyungho",
editor="Li, Lei
and Jirsa, Viktor
and Feng, Jianfeng
and Deng, Jun
and Dede', Luca
and An, Sora
and Lyu, Yilin
and Liu, Xiaoyue",
title="Towards Digital Twin of {RF} Ablation: Real-Time Prediction of Time-Dependent Thermal Effects Using Transformer",
booktitle="Digital Twin for Healthcare",
year="2026",
publisher="Springer Nature Switzerland",
address="Cham",
pages="69--78",
isbn="978-3-032-07694-6"
}

\end{document}